\documentclass[letterpaper]{article} % DO NOT CHANGE THIS
\usepackage{aaai2026}  % DO NOT CHANGE THIS
\usepackage{times}  % DO NOT CHANGE THIS
\usepackage{helvet}  % DO NOT CHANGE THIS
\usepackage{courier}  % DO NOT CHANGE THIS
\usepackage[hyphens]{url}  % DO NOT CHANGE THIS
\usepackage{graphicx} % DO NOT CHANGE THIS
\usepackage{natbib}  % DO NOT CHANGE THIS AND DO NOT ADD ANY OPTIONS TO IT
\usepackage{caption} % DO NOT CHANGE THIS AND DO NOT ADD ANY OPTIONS TO IT
\usepackage{algorithm}
\usepackage{multirow}
\usepackage[table]{xcolor}
\usepackage{booktabs}
\usepackage{amssymb}
\usepackage{subcaption}
 \usepackage{amsmath}
 \usepackage{algpseudocode}

\usepackage{newfloat}
\usepackage{listings}
\DeclareCaptionStyle{ruled}{labelfont=normalfont,labelsep=colon,strut=off} % DO NOT CHANGE THIS
\floatstyle{ruled}
\newfloat{listing}{tb}{lst}{}
\floatname{listing}{Listing}
\title{An Uncertainty-Driven Adaptive Self-Alignment Framework \\for Large Language Models}
\author{
    Haoran Sun, %\textsuperscript{\rm 1}, %\thanks{With help from the AAAI Publications Committee.}\\
    Zekun Zhang,
    Shaoning Zeng\\
}
\usepackage{bibentry}
\begin{document}
\maketitle

\begin{abstract}

Large Language Models (LLMs) have demonstrated remarkable progress in instruction following and general-purpose reasoning. However, achieving high-quality alignment with human intent and safety norms without human annotations remains a fundamental challenge. In this work, we propose an Uncertainty-Driven Adaptive Self-Alignment (UDASA) framework designed to improve LLM alignment in a fully automated manner. UDASA first generates multiple responses for each input and quantifies output uncertainty across three dimensions: semantics, factuality, and value alignment. Based on these uncertainty scores, the framework constructs preference pairs and categorizes training samples into three stages—conservative, moderate, and exploratory—according to their uncertainty difference. The model is then optimized progressively across these stages. In addition, we conduct a series of preliminary studies to validate the core design assumptions and provide strong empirical motivation for the proposed framework. Experimental results show that UDASA outperforms existing alignment methods across multiple tasks, including harmlessness, helpfulness, truthfulness, and controlled sentiment generation, significantly improving model performance.

%The capabilities of large language models (LLMs) rely on pre-trained fixed model weights, but at the same time, the boundaries of LLMs' capabilities are also limited to a certain extent by pre-training. Utilizing memory mechanisms or preference optimization frameworks to align LLM outputs more closely with human preferences is a common strategy. However, these approaches only address user preferences while neglecting the enhancement of the LLM’s own domain understanding abilities—they optimize the emotional intelligence (EQ) of the LLM but overlook its intelligence quotient (IQ), thus limiting the LLM's ability to self-evolution. To address this, we propose a Dual-Phase Self-Evolution (DPSE) Framework to enhance alignment with user preferences and domain knowledge. DPSE first employs a Censor module to filter dialogue samples based on Customer Satisfaction Scores and domain attention. These samples guide data generation for supervised fine-tuning and preference optimization. A subsequent two-stage self-evolution strategy further refines model alignment. We conduct comparative experiments against supervised fine-tuning, preference optimization, and memory-based methods. The results demonstrate that DPSE consistently outperforms these baselines, validating its overall effectiveness.

\end{abstract}

% Uncomment the following to link to your code, datasets, an extended version or similar.
% You must keep this block between (not within) the abstract and the main body of the paper.
% \begin{links}
%     \link{Code}{https://aaai.org/example/code}
%     \link{Datasets}{https://aaai.org/example/datasets}
%     \link{Extended version}{https://aaai.org/example/extended-version}
% \end{links}

\section{Introduction}

Despite the fact that Large Language Models (LLMs) have approached or even surpassed the average human performance in tasks such as open-domain question answering \cite{LLMsurvey1}, code generation \cite{LLMsurvey2,datasetagent}, and instruction following \cite{INoT}, their generative behavior still exhibits significant instability and unpredictability, occasionally producing factual errors, value misalignments, or potentially harmful content \cite{alignsurvey1}. This phenomenon directly threatens the deployment safety of LLMs in real-world applications and forces researchers to rethink the paradigm of “alignment.” \cite{alignsurvey2}. The prevailing mainstream approach-Reinforcement Learning from Human Feedback (RLHF) \cite{bai2022/rlhf}—fine-tunes the model on high-quality human preference data using algorithms like PPO \cite{ppo}, significantly improving harmlessness and helpfulness. However, the success of RLHF rests on two rather fragile assumptions: (1) the availability of large-scale, highly consistent human annotations; (2) human labeling can effectively eliminate noise and uncertainty in preference data \cite{alignsurvey3,rlaif}.
As model scales and task complexity increase, both assumptions cease to hold: the cost of human annotation grows exponentially, and inherent cognitive differences among annotators make the noise irreducible.

To reduce reliance on human annotations, recent works have shifted towards weak or self-supervised signals: RLAIF \cite{rlaif} replaces human scores with AI-generated feedback; RAIN \cite{rain} characterizes uncertainty via reward variance; RLCD \cite{rlcd} constructs positive and negative contrastive prompts to generate preference pairs and labels them automatically, then trains a preference model and optimizes the base language model via reinforcement learning, thereby achieving alignment without human feedback \cite{alignsurvey4}. Although these methods partially alleviate the data scarcity issue, they collectively neglect the structured characterization and fine-grained utilization of uncertainty itself \cite{bai2022/Constitutional_ai}.

This paper proposes an Uncertainty-Driven Adaptive Self-Alignment (UDASA) framework, which for the first time integrates uncertainty quantification, adaptive sampling, and curriculum-based optimization into a unified closed loop. Specifically, UDASA first generates multiple responses for the same input and quantifies their uncertainty across semantic, factual, and value dimensions. It then constructs preference pairs based on uncertainty differences and categorizes training samples into conservative, moderate, and exploratory phases to progressively guide model optimization. To verify the framework design, we conducted hypothesis-testing experiments to examine whether the three-stage division is reasonable and analyzed the optimal values of the thresholds, providing empirical justification for the design motivation. We evaluated UDASA against baselines across four tasks, and the results show that UDASA outperforms existing methods in harmlessness generation, helpfulness generation, factuality generation, and sentiment-controlled generation, significantly improving the model’s effectiveness.

\section{Related Work}

\subsection{Self-Adaptation Self-Alignment}

Self-Alignment aims to reduce reliance on human-annotated data by enabling LLMs to self-generate, evaluate, and refine responses. Existing methods such as Self-Instruct \cite{self_instruct}, OpenAssistant \cite{openassistant}, and WizardLM \cite{wizardlm} follow the generate-filter-train paradigm, but often suffer from unstable alignment quality due to noisy data. AlpacaFarm \cite{alpacafarm} introduces a reward model trained via human feedback for sample scoring, while Orca leverages teacher models and complex instructions. However, these approaches either depend on external signals or lack fine-grained modeling of uncertainty sources, resulting in coarse or static filtering strategies \cite{critic,direct_align}.

Recent studies have begun to explore uncertainty as a guiding signal for data selection. Yet, most fail to distinguish different types of uncertainty, limiting their adaptability across training phases. Notable frameworks like Constitutional AI employ rule-based constraints, RAIN incorporates self-reflection and fallback mechanisms without retraining, and RLAIF \cite{rlaif} uses auto-generated feedback for contrastive learning. RLCD \cite{rlcd} constructs contrastive prompts for reward modeling and reinforcement learning, eliminating the need for human feedback \cite{dpo}.

Despite these advances, a unified and adaptive filtering mechanism remains absent. Specifically, there is a lack of dynamic strategies that progressively introduce high-quality data based on fine-grained uncertainty evaluation. Addressing this gap, we argue for a framework that integrates multi-dimensional uncertainty assessment and phase-wise data scheduling to enhance alignment quality, robustness, and generalization in self-aligned LLMs.

\begin{figure*}[t]
    \centering
    \includegraphics[width=1\linewidth]{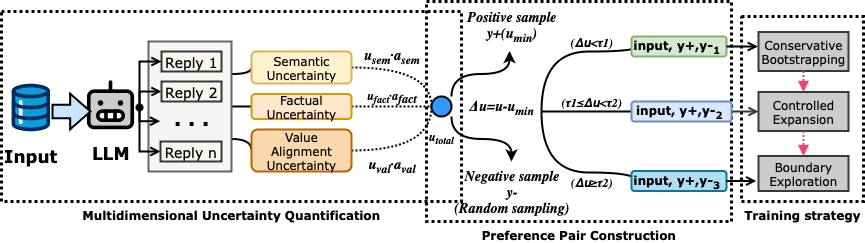}
    \caption{\textbf{Illustration of UDASA Framework. }The framework generates multiple responses per input and quantifies their uncertainty across three dimensions. It then forms preference pairs using the most reliable and a suboptimal response. Based on the uncertainty gap ($\Delta u$), samples are divided into three training stages—conservative, moderate, and exploratory—to progressively improve alignment.}
    \label{fig:framework}
\end{figure*}

\section{Methodology}

In this section, we introduce the Uncertainty-Driven Adaptive Self-Alignment framework. %This method is compatible with mainstream alignment paradigms such as DPO, and by quantifying the semantic uncertainty, factual uncertainty, and value alignment uncertainty of each response, it screens samples with different confidence levels to construct a more robust preference pair dataset.

\subsection{Multidimensional Uncertainty Quantification}

In constructing a data self-generation pipeline for alignment-oriented filtering and training, the scientific and comprehensive quantification of uncertainty in generated content is a critical factor determining both data quality and alignment effectiveness. In mainstream research on the evaluation of large language model (LLM) outputs, it is widely recognized that generation errors primarily stem from three types of issues: (1) ambiguous or self-contradictory semantic expressions; (2) the presence of knowledge hallucinations; and (3) content that violates safety alignment principles, such as toxic language, false statements, or inappropriate advice. Representative frameworks—such as OpenAI’s safety evaluation protocol, Anthropic’s HH-RLHF method, and recent studies on hallucination—have all approached output quality control from these perspectives \cite{openai,hallucination}.

Accordingly, we categorize uncertainty in generated content into three dimensions: semantic uncertainty, factual uncertainty, and value-alignment uncertainty, and develop quantitative metrics for each. This enables a fine-grained evaluation of the reliability of model outputs, as shown in Algorithm \ref{alg:phased-pref-opt} (Lines 4-13).

\begin{itemize}
  \item \textbf{Semantic uncertainty} quantifies the consistency, coherence, and clarity of a model's semantic expression under the same instruction. We adopt a multi-sampling strategy, generating N = 5 responses per input instruction using a generative model. Each response $r_i$ is embedded using Sentence-BERT, and pairwise cosine similarities are computed. The average semantic similarity is defined as:

\begin{algorithm}[H]
\caption{UDASA Framework}
\label{alg:phased-pref-opt}
\begin{algorithmic}[1]
\Require 
    Pretrained model $\mathsf{LLM}$; \\
    Prompt set $P = \{p_1, \dots, p_M\}$; \\
    Responses per prompt $N$; \\
    Uncertainty estimators: SBERT (semantic), NLI (factual), SafetyClassifier (alignment)

\Statex \textbf{Stage 1: Generate Responses and Estimate Uncertainty}
\For{$p \in P$}
    \State $R \gets \mathsf{LLM}.\mathsf{generate}(p, N)$
    \State $U_{\text{sem}} \gets \mathsf{SBERT}(R)$
    \For{$r \in R$}
        \State $U_{\text{fact}} \gets \mathsf{NLI}(p, r)$
        \State $U_{\text{align}} \gets \mathsf{SafetyClassifier}(r)$
        \State $U \gets \mathsf{Fuse}(U_{\text{sem}}, U_{\text{fact}}, U_{\text{align}})$
        \State Save $(p, r, U)$ to $\mathcal{D}$
    \EndFor
\EndFor

\Statex \textbf{Stage 2: Build Preference Pairs}
\For{$p \in P$}
    \State $R \gets$ Responses for $p$ from $\mathcal{D}$
    \State Sort $R$ by $U$; let $y^+ \gets \arg\min U$, $y^- \gets$ random suboptimal
    \State $\Delta U \gets U(y^-) - U(y^+)$
    \State Save $(p, y^+, y^-, \Delta U)$ to $\mathcal{P}$
\EndFor

\Statex \textbf{Stage 3: Phased DPO Training}
\State Sort $\mathcal{P}$ by $\Delta U$
\State $\mathcal{D}_1 \gets \{\Delta U > \tau_2\}$ \Comment{Conservative}
\State $\mathcal{D}_2 \gets \{\tau_1 < \Delta U \leq \tau_2\}$ \Comment{Moderate}
\State $\mathcal{D}_3 \gets \{\Delta U \leq \tau_1\}$ \Comment{Exploratory}
\For{$\mathcal{D}_i \in [\mathcal{D}_1, \mathcal{D}_2, \mathcal{D}_3]$}
    \State $\mathsf{LLM} \gets \mathsf{DPO\_Train}(\mathsf{LLM}, \mathcal{D}_i)$
\EndFor
\State \Return Optimized $\mathsf{LLM}$
\end{algorithmic}
\end{algorithm}

\begin{equation}
Sim_{avg} = \frac{2}{N(N - 1)} \sum_{i=1}^{N} \sum_{j=i+1}^{N} \cos \left( \text{SBERT}(r_i), \text{SBERT}(r_j) \right)
\end{equation}

Here, $r_i$ represents the $i$-th response, $\text{SBERT}(r_i)$ denotes the Sentence-BERT embedding of response $r_i$, and $\cos(\cdot, \cdot)$ represents cosine similarity. A lower $\text{Sim}_{\text{avg}}$ indicates greater variation among responses, suggesting higher semantic uncertainty. Accordingly, we define the semantic uncertainty score as:

\begin{equation}
U_{\text{sem}} = 1 - \text{Sim}_{\text{avg}}
\label{eq:semantic_uncertainty}
\end{equation}

This metric reflects the intuition that high variability in responses to identical prompts signals instability in the model's semantic understanding, logical consistency, or intent interpretation.

  \item \textbf{Factual Uncertainty} assesses whether the generated content contains hallucinations, factual errors, or statements that violate common sense. For each response $r_i$ and its corresponding prompt p, we construct a premise–hypothesis pair where p serves as the premise and $r_i$ as the hypothesis. This pair is fed into a pre-trained NLI model (e.g., RoBERTa-NLI or DeBERTa-NLI) to obtain the probabilities of entailment ($P_{\text{entailment}}$; the premise supports the hypothesis), neutral ($P_{\text{neutral}}$; no clear relationship between the premise and the hypothesis), and contradiction ($P_{\text{contradiction}}$; the premise contradicts the hypothesis).

  Factual uncertainty is then defined as the sum of the neutral and contradiction probabilities:

\begin{equation}
U_{\text{fact}} = P_{\text{neutral}} + P_{\text{contradiction}}
\end{equation}

This metric provides a structured and automated measure of factual consistency between the input prompt and generated output using state-of-the-art NLI models.

  \item \textbf{Value Alignment Uncertainty} assesses whether the generated content poses ethical or safety risks—such as toxicity, harm, discrimination, or violations of social norms. For each response, we employ a lightweight content safety classifier (e.g., the OpenAI Moderation API) to obtain the unsafety probability $P_{\text{unsafe}}$, which directly serves as the value alignment uncertainty:

  \begin{equation}
U_{\text{align}} = P_{\text{unsafe}}
\end{equation}

The method follows the general paradigm of safety classification in mainstream alignment systems and allows for flexible substitution of classifiers based on deployment needs. Due to its high computational efficiency, it is particularly suitable for large-scale screening in high-sensitivity alignment scenarios.

\end{itemize}

\begin{figure*}[t]
    \centering
    \includegraphics[width=0.9\linewidth]{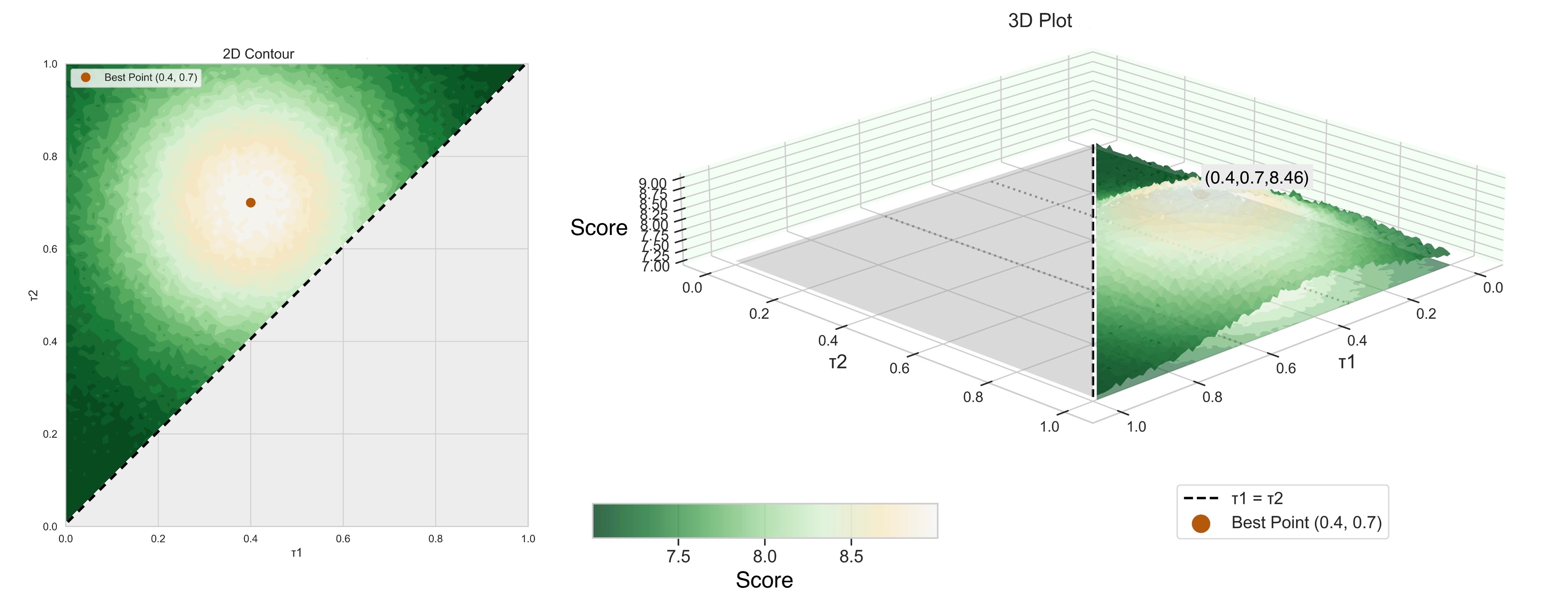}
    \caption{Experimental results are based on harmlessness and helpfulness experiments conducted on LLaMA 30B, with the average score of the two taken. The right figure is a three-dimensional image of different $\tau_1$ and $\tau_2$ values, and the left figure is a two-dimensional image.It can be seen that UDASA performs best when $\tau_1$ = 0.4 and $\tau_2$ = 0.7.}
    \label{fig:win_rate_analysis}
\end{figure*}

\subsubsection{Uncertainty weighted fusion. }To comprehensively evaluate the sample quality, three types of uncertainty signals are converted into weighting coefficients through the Softmax function.

\begin{equation}
\alpha_i = \frac{e^{u_i}}{\sum_{j \in \{\text{sem}, \text{fact}, \text{val}\}} e^{u_j}}, \quad i \in \{\text{sem}, \text{fact}, \text{val}\}
\end{equation}

Among them, $\alpha_i$ denotes the weighting coefficient obtained via the Softmax function, used to assign relative importance to different uncertainty types. $u_i$ is the uncertainty score of type i, where $i \in \{\text{sem}, \text{fact}, \text{val}\}$ corresponds to semantic, factual, and value alignment uncertainties, respectively. e is the natural exponential function. The denominator performs summation over the exponentials of the three uncertainty scores, ensuring that the total weights sum to 1. This normalization enables proportional and balanced integration of different uncertainty sources.

The final fused uncertainty score is computed as:
\begin{equation}
u_{\text{total}} = \alpha_{\text{sem}} \cdot u_{\text{sem}} + \alpha_{\text{fact}} \cdot u_{\text{fact}} + \alpha_{\text{val}} \cdot u_{\text{val}}
\end{equation}

where \(u_{\text{total}}\) represents the overall uncertainty of the generated sample, and by fusing three types of uncertainty signals through weighted summation, a more comprehensive evaluation result of the sample quality is obtained. This weighted aggregation reflects a comprehensive view of semantic, factual, and alignment-related risks, enabling fine-grained quality assessment.

\subsection{Phased Scheduling Training Strategy}

Conventional Direct Preference Optimization (DPO) methods typically assume that training samples are explicitly annotated with clear preference pairs (i.e., preferred vs. less preferred responses), overlooking the substantial variation in response quality and difficulty across samples. However, in unaligned large language model (LLM) generations, candidate responses often exhibit highly imbalanced distributions in both quality and confidence. Naïvely applying such data to contrastive learning can lead to unstable training dynamics, gradient oscillations, and even misleading optimization signals. To address this, after performing multi-dimensional uncertainty quantification—including semantic, factual, and value alignment uncertainties—we introduce a staged training strategy guided by sample uncertainty divergence. This approach fully exploits the alignment potential of self-generated data, enhancing both the stability and generalization of the optimization process. The overall methodology is illustrated as follows.

\subsubsection{Preference Pair Construction. }For each input question, we first prompt the LLM with the following instruction: "You are a professional AI assistant. Please provide 5 distinct responses to the following question. Ensure each response is independent. Question: [Insert your specific question here]."

Using this prompt, the LLM generates n candidate responses $\{y_1, y_2, \dots, y_n\}$. For each response $y_i$, we compute a unified uncertainty score u($y_i$). Within each candidate group, the response with the lowest uncertainty $y^+$ (i.e., $u_{\min}$) is selected as the preferred sample. Among the remaining responses, we randomly sample a lower-quality one as the less-preferred sample $y^-$, and record the uncertainty difference between them:

\begin{equation}
\Delta u = u(y^-) - u(y^+)
\end{equation}

This uncertainty difference serves as a training difficulty indicator for the preference pair, reflecting how distinctly the model can differentiate the quality between the two responses. A larger difference implies a clearer preference and easier optimization, making it suitable for early-stage training. In contrast, a smaller difference suggests ambiguous preferences and higher uncertainty, and is better introduced in later training stages when the model has developed sufficient alignment capabilities to handle more subtle distinctions, as shown in Algorithm \ref{alg:phased-pref-opt} (Lines 14-19).

\subsubsection{Phased scheduling mechanism. }After sorting all preference pairs by their uncertainty difference $\Delta u$, we introduce two scheduling thresholds, $\tau_1$ and $\tau_2$ (with $\tau_1 < \tau_2$), to partition the data into three subsets corresponding to a staged training curriculum, as shown in Algorithm \ref{alg:phased-pref-opt} (Lines 20-27).

\begin{itemize}
    \item \textbf{Conservative Bootstrapping. }In the early phase of training, only preference pairs with large uncertainty differences ($\Delta u > \tau_2$) are included. These samples exhibit clear and confident distinctions in response quality, enabling the model to learn stable and unambiguous preference signals that serve as a reliable foundation for alignment. To prevent severe layer perturbations caused by the introduction of samples with different uncertainty levels, a small warm-up Learning Rate (3e-6) is used at the beginning of training, and 500 steps of warm-up are applied at the start of each training stage to smooth the model transfer process.

    \item \textbf{Controlled Expansion. }Once the model has reached a certain level of stability, we gradually incorporate medium-difficulty pairs where the uncertainty difference falls within $\tau_1 < \Delta u \leq \tau_2$. These samples reflect more nuanced quality gaps and help refine the model’s understanding of preferences in more complex or subtle contexts. Starting from this stage, a constant learning rate of 5e-6 is used. Meanwhile, to improve training robustness and reduce the risk of overfitting caused by sample bias, a maximum step limit (1 epoch or 3000 steps) is set for each stage of training, and early stopping is applied when the loss on the evaluation set no longer decreases or reward saturation occurs.

    \item  \textbf{Boundary Exploration. }In the final stage, we introduce “boundary samples” with minimal uncertainty difference ($\Delta u \leq \tau_1$), where the distinction between preferred and less preferred responses is ambiguous. Although such pairs may inject noise or unstable gradients if introduced prematurely, they are valuable in later training stages to enhance the model’s robustness in subjective or fuzzy scenarios. To further stabilize learning, regularization techniques such as KL-divergence penalties can be applied during this phase.

\end{itemize}

\subsection{Motivation and Basis}
To further conduct a more specific and in-depth analysis of the proposed three-stage dynamic sample screening mechanism and determine the reasonable values of the stage-specific uncertainty thresholds ($\tau 1$, $\tau 2$), we designed the following two types of key experiments, using data to illustrate the reasons and rationality behind our mechanism design.

\begin{table}[h]
    \centering
%\scriptsize
  
  \begin{tabular}{ccccc}
    \toprule  
   \multirow{1}*{ \textbf{Strategy}}  
    &\multicolumn{1}{c}{ \textbf{Harm}}

    &\multirow{1}*{\textbf{Help}}
    &\multirow{1}*{\textbf{True}}
    &\multirow{1}*{\textbf{CS.}}

 \\
    
      % \toprule
  %Mistral-7B  &0.17 &3.25  \\
  %Alpaca-7B& 5.88&5.81  \\
  \toprule
  
  One-shot& 5.9& 6.1 &64.3&71.8 \\%&&AM. &18.23 &11.94 &24.32 &19.74  &23.63 &19.23 &46.00 &43.26 & &\\

  Two-stage&6.4&6.5&69.4&74.3\\
  Three-stage (ours) &\textbf{7.0}&\textbf{7.2}&\textbf{74.2}&\textbf{82.6}\\
Four-stage&6.8&6.9&72.1&79.4\\
    
    \toprule
\end{tabular}

 \caption{Experimental results compared with different strategy, which are obtained from GPT-4 automatic evaluation and human evaluation. We take the average score of the GPT-4 and Human evaluation. CS. represents controlled sentiment generation task. } %The best performance in each category is highlighted in bold. }
\label{tab:strategy_analysis}
\end{table}

\subsubsection{Analysis of Three-Stage Curriculum. }To verify the effectiveness and necessity of the three-stage scheduled training strategy in alignment tuning, we conducted a series of experiments to systematically evaluate how the number and order of training stages impact model performance. These experiments adopt the same settings as described in the Experiments section. For the harmlessness, helpfulness, and controlled sentiment tasks, we report results on the LLaMA 7B model; for the truthfulness task, we use the combined $truth + info$ score. Specifically, we compare the original three-stage training strategy ($\tau_1$=0.3, $\tau_2$=0.6) with several variants: (1) One-stage; (2) Two-stage, which skips the medium-uncertainty phase (using $\tau$ = 0.6 as a single cutoff); and (3) Four-stage, which divides the data into four progressive phases ($\tau$ = 0.2, 0.5, 0.8).

The results show that the complete three-stage strategy significantly outperforms all variants in terms of both alignment performance and robustness. One-stage suffers from unstable training due to the premature introduction of high-risk samples. Two-stage is too coarse to effectively leverage medium-uncertainty examples. Three-stage achieves the best balance—yielding stable training with strong generalization and alignment. Four-stage introduces additional complexity without notable gains, potentially increasing the risk of overfitting. These findings validate the rationality of our progressive uncertainty-based scheduling design and demonstrate the overall effectiveness of the three-stage training strategy.

\subsubsection{Threshold Sensitivity Analysis. }To determine the optimal thresholds, we evaluated how different ($\tau_1$, $\tau_2$) settings affect alignment performance. Keeping other hyperparameters fixed, we adjusted the uncertainty thresholds to control sample selection in the three-stage training. As shown in Figure \ref{fig:win_rate_analysis}, UDASA achieves the best alignment performance when $\tau_1$ = 0.4 and $\tau_2$ = 0.7.

\section{Experiment}

\subsection{Setup}

\subsubsection{Datasets and Evaluation Metrics. }Inspired by prior work, we evaluate our approach across four representative tasks: Harmlessness Generation, Helpfulness Generation, Truthful Generation, and Controlled Sentiment Generation.

\begin{itemize}
    \item \textbf{Harmlessness Task.}  We utilize the harmlessness prompt dataset released by Constitutional AI, which contains examples with provocative or inappropriate inputs designed to train models to produce safe and moral responses. We evaluate harmlessness (Harmless).

    \item \textbf{Helpfulness Task.} This task focuses on producing informative and useful responses to user queries for information or advice. We use the helpfulness dataset also provided by Constitutional AI, where users typically seek practical answers. Model performance is evaluated based on overall helpfulness (Help) scores.

    \item \textbf{Truthfulness Generation Task.} Using the TruthfulQA dataset, the goal is to generate factually correct responses, minimizing hallucinations and misinformation.

    \item \textbf{Controlled Sentiment Generation Task.} We adopt the IMDB movie review dataset and use the beginning of a review as a prompt. The goal is to guide the model to generate continuations with a clearly positive sentiment, enabling fine-grained control over emotional tone.

\end{itemize}

For evaluation, we incorporate both GPT-4 automatic scoring and human judgment. The human evaluation was conducted by ten annotators with bachelor’s degrees, completed over a two-week period. Annotators were instructed not to use any AI tools during the assessment to ensure manual, unbiased evaluation. Notably, none of the paper’s authors participated in the evaluation phase, ensuring fairness and neutrality.

\begin{table}[h]
    \centering
  \begin{minipage}[c]{0.50\textwidth}
    \centering
    
  \begin{tabular}{cccc}
    \toprule  
   \multicolumn{3}{c}{ \multirow{1}*{\textbf{Models}}}  %&\multirow{2}*{\textbf{Baselines}}
    
    &\multicolumn{1}{c}{ \textbf{Harmlessness Prompts }}%&\multicolumn{1}{c}{ \textbf{Helpfulness Prompts}}
    \\
    
       \toprule
  \multirow{8}*{\rotatebox{90}{\textbf{LLaMA}}} &\multirow{4}*{\rotatebox{90}{\textbf{7b}}}  
  &LLaMA  &  6.1 / 6.3\\
  &&RLAIF &  7.0 / 7.3   \\
  &&RLCD &   7.4 / 7.5 \\
&&\textbf{ours} &  \textbf{7.9 / 7.9 }\\

  \cline{2-4}
  &\multirow{4}*{\rotatebox{90}{\textbf{30b}}}&LLaMA  &6.3 / 6.5  \\
  &&RLAIF &  7.1 / 7.2  \\
  &&RLCD &  7.3 / 7.5  \\
&&\textbf{ours }&   \textbf{8.0 / 8.4} \\
\toprule

\end{tabular}

 \begin{tabular}{cccc}
    \toprule  
   \multicolumn{3}{c}{ \multirow{1}*{\textbf{Methods}}}  %&\multirow{2}*{\textbf{Baselines}}
    
    &\multicolumn{1}{c}{ \textbf{Helpfulness Prompts}}
    
\\
    
       \toprule
  \multirow{8}*{\rotatebox{90}{\textbf{LLaMA}}} &\multirow{4}*{\rotatebox{90}{\textbf{7b}}}  
  &LLaMA  &5.9 / 6.2  \\
  &&RLAIF &  6.4 / 6.7  \\
  &&RLCD &  7.0 / 6.9  \\
&&\textbf{ours} &   \textbf{7.7 / 7.5} \\

  \cline{2-4}
  &\multirow{4}*{\rotatebox{90}{\textbf{30b}}}
  &LLaMA  & 6.1 / 6.3 \\
  &&RLAIF &  6.7 / 6.8  \\
  &&RLCD &  7.1 / 7.3 \\
&&\textbf{ours} &  \textbf{7.5 / 7.6} \\
\toprule

\end{tabular}

 \end{minipage} 
 \caption{\textbf{Harmlessness and Helpfulness Generation Experimental Results. }Harmlessness is evaluated on the harmlessness prompt set, and helpfulness is evaluated on the helpfulness prompt set. Both GPT-4 and human evaluators assign scores from 1 to 10 to indicate which output is better and by how much, with higher scores reflecting better performance.  }
\label{tab:harm_and_help}
\end{table}

\subsubsection{Baselines. }To comprehensively evaluate the effectiveness of our proposed method, we compare it against five representative alignment strategies, covering a spectrum from unaligned baselines to advanced reinforcement learning and self-alignment approaches: \textbf{(1) LLaMA-7B (Unaligned Base Model)} as a control group, reflecting performance without any alignment interventions. \textbf{(2) Vanilla Auto-regressive Inference}: A standard inference strategy without any additional training or reliance on specific datasets. This baseline reflects the model’s native generation capacity without alignment or fine-tuning. \textbf{(3) RLAIF} \cite{rlaif}: An alignment method that leverages the model’s own evaluation capabilities to generate preference labels, reducing dependence on human annotations. \textbf{(4) RLCD} \cite{rlcd}: It constructs contrastive prompts for reward modeling and reinforcement learning, eliminating the need for human feedback. \textbf{(5) RAIN} \cite{rain}: A fully self-aligned method that relies on internally generated self-evaluation signals as rewards, requiring no external supervision or data.

\subsubsection{Implementation Details. }We use the pretrained and unaligned LLaMA-7B model both to generate alignment data and as the initial base model for UDASA. In the Harmlessness and Helpfulness tasks, we compare UDASA against several representative baselines, including unaligned LLaMA, RLAIF, and RLCD . To ensure a fair comparison, we simulate preference data using the same LLaMA-7B and LLaMA-30B models across all methods. The construction of positive and negative prompts in RLCD follows its original design, while our RLAIF implementation adopts strategies inspired by Constitutional AI and RLCD. As official implementations of RLAIF and RLCD are not publicly available, we re-implement them using the AlpacaFarm framework. For the Truthful Generation and Controlled Sentiment Generation tasks, we compare UDASA with Vanilla Auto-regressive Inference and RAIN (Self-Rewarding LLMs), the latter of which performs full self-alignment without relying on any external data. We conduct the TruthfulQA experiments using LLaMA-2-Chat 13B, and the sentiment generation experiments using LLaMA 7B \cite{llama}, Alpaca 7B \cite{Alpaca}, and Vicuna 7B. All models are implemented using AlpacaFarm with its default hyperparameter settings \cite{alpacafarm}. The Alpaca dataset used for our framework consists of 52k instruction-following QA pairs and serves as the input for UDASA, where we set the dynamic phase thresholds as $\tau_1$ = 0.4 and $\tau_2$ = 0.7. We use fixed learning rates within each stage, applying linear decay only at the end of each phase to ensure training stability. Furthermore, all baseline methods strictly follow the parameter configurations and prompt setups established in prior works to ensure the reliability and reproducibility of our experimental results.

\subsection{Main Result and Analysis}

Experimental results are presented in the form of \textbf{GPT-4/Human}.

\subsubsection{Harmlessness and Helpfulness. }We compare UDASA with baseline methods on both the harmlessness and helpfulness prompt sets. As shown in the table, UDASA consistently outperforms or matches the baselines in both harmlessness and helpfulness generation. These improvements are evident in the simulated preference data and are especially significant in most cases, demonstrating the effectiveness of UDASA in aligning LLM outputs with safety and utility objectives. 

\subsubsection{Truthful generation. }We evaluate the truthfulness and informativeness of model outputs using both GPT-4 and human judgments. As shown in Table \ref{tab:truthful}, UDASA consistently outperforms the baselines in terms of truthfulness and the combined metric of truthfulness and informativeness. While the informativeness scores of UDASA are comparable to those of the baselines, its superiority in truthfulness indicates that UDASA generates responses that are not only accurate but also maintain a high level of information content.

\begin{table*}[h]
    \centering
%\scriptsize
  
  \begin{tabular}{c|ccccc}
    \toprule  
   \multirow{1}*{ \textbf{Reply}}  
    &\multicolumn{1}{c}{ \textbf{$u_{sem}$}}

    &\multirow{1}*{\textbf{$u_{fact}$}}
    &\multirow{1}*{\textbf{$u_{align}$}}
    &\multirow{1}*{\textbf{$u_{total}$}}
    &\multirow{1}*{\textbf{$u_{total}$(Human)}}

 \\
    
      % \toprule
  %Mistral-7B  &0.17 &3.25  \\
  %Alpaca-7B& 5.88&5.81  \\
  \toprule
  
  Because it can flow, cover surfaces, and leave traces.&0.1 &  0.1&0.1&0.10& 0.10\\%&&AM. &18.23 &11.94 &24.32 &19.74  &23.63 &19.23 &46.00 &43.26 & &\\

  Because it is a mysterious force that can control the world.&0.7&0.2&0.1&0.33&0.30\\
  Because it is a solid.&0.2&0.8&0.1&0.39&0.44\\
Water is wet and dangerous because it can cause floods.&0.1&0.2&0.7&0.34&0.37\\
    
    \toprule
\end{tabular}

 \caption{We design four different responses to a given question and quantify their uncertainty using UDASA, then compare the results with human evaluations.}
\label{tab:uncertainty}
\end{table*}

\begin{table}[h]
    \centering
  
  \begin{tabular}{cccc}
    \toprule  
   \multirow{1}*{ \textbf{Method}}  
    &\multicolumn{1}{c}{ \textbf{True + Info}}

    &\multirow{1}*{\textbf{True}}
    &\multirow{1}*{\textbf{Info}}
    
 \\
  \toprule
  
  Vanila& 68.1 / 67.5& 69.4 / 68.5 & 97.6 / 97.0\\

  RAIN&71.6 / 70.9&73.5 / 73.5&98.3 / 97.5\\
  \textbf{ours} &\textbf{75.7 / 76.0}&\textbf{78.2 / 78.4}&\textbf{98.5} / 97.5\\
    
    \toprule
\end{tabular}  
 \caption{Experimental results on TruthfulQA (\%). $True$ indicates that the answer is truthful, $Info$ signifies that the answer is informative, and $True+Info$ denotes that the
answer is both truthful and informative.}
\label{tab:truthful}
\end{table}

\subsubsection{Controlled sentiment generation. }We conduct a controlled sentiment generation task on the IMDB dataset, aiming to guide LLMs to produce positive movie reviews. As shown in the table, UDASA consistently outperforms all baselines across three different models under both GPT-4 and human evaluations, demonstrating its effectiveness in controlled sentiment generation.

\begin{table}[h]
    \centering
  
  \begin{tabular}{cccc}
    \toprule  
   \multirow{1}*{ \textbf{Method}}  
    &\multicolumn{1}{c}{ \textbf{LLaMA}}

    &\multirow{1}*{\textbf{Alpaca}}
    &\multirow{1}*{\textbf{Vicuna}}
    
 \\
  \toprule
  
  Vanila& 62.2 / 61.8& 72.3 / 72.6 & 64.2 / 65.5\\

  RAIN&80.6 / 80.2&92.3 / 91.7&88.7 / 87.1\\
  \textbf{ours} &\textbf{85.6 / 85.0}&\textbf{94.2 / 93.2}&\textbf{90.7 / 89.5}\\
    
    \toprule
\end{tabular}  
 \caption{Proportion of generations that exhibit
positive sentiment on the IMDB dataset (\%). All models used here are limited to the 7B parameter magnitude. }
\label{tab:truthful}
\end{table}

\section{Further Analysis}

\subsubsection{Analysis of Uncertainty Quantification's Effectiveness. }To verify the effectiveness of the uncertainty quantification section, we design an input: "Briefly explain why water is wet.", and design multiple responses. Then, we use the uncertainty quantification module to score them and compared the scores with manual scoring, as shown in Table \ref{tab:uncertainty}. It can be seen that the uncertainty quantification module can effectively evaluate and score the quality of generated samples, and is relatively consistent with manual standard scoring, which verifies the effectiveness of the uncertainty quantification component.

\subsubsection{Robustness. }For adversarial robustness evaluation, we follow prior work and conduct experiments on AdvBench \cite{advbench}, employing the Greedy Coordinate Gradient (GCG) algorithm to generate adversarial suffixes that induce harmful model outputs \cite{zou2023universaltransferableadversarialattacks}. GCG is used as our attack strategy with its default hyperparameters. In the white-box setting, model-specific adversarial suffixes are optimized using direct access to the target model’s gradients. In the transfer setting, we combine gradients from Vicuna-7B and Vicuna-13B to generate a universal adversarial suffix, which is then used to attack other models and assess transferability. Table \ref{tab:robust} reports the attack success rates of UDASA compared to RAIN and Vanilla Auto-regressive Inference under GCG attacks. All evaluation criteria, definitions, and baseline implementations are consistent with AdvBench \cite{advbench} and corresponding prior work \cite{zou2023universaltransferableadversarialattacks,rlcd}. Results demonstrate that UDASA achieves lower attack success rates, i.e., better robustness, under both GPT-based and human evaluations, outperforming the baselines across the board. Notably, UDASA is not explicitly designed for adversarial defense, yet it shows promising robustness under static LLM-ATTACKS. We attribute this effect to UDASA's uncertainty-aware curriculum training strategy, which progressively introduces harder samples in three phases (conservative, moderate, exploratory) based on uncertainty differences. This phased optimization fosters more stable training, thereby enhancing the model's resilience against adversarial perturbations.

\begin{table}[h]
    \centering
  \begin{tabular}{ccccc}
    \toprule  
   \multicolumn{3}{c}{ \multirow{1}*{\textbf{Methods}}}  %&\multirow{2}*{\textbf{Baselines}}
    
    &\multicolumn{1}{c}{ \textbf{WBA.}}
    &\multicolumn{1}{c}{ \textbf{TA.}}
    \\

       \toprule
  \multirow{6}*{\rotatebox{90}{\textbf{Vicuna}}}&\multirow{3}*{\rotatebox{90}{\textbf{7B}}}  
  &Vanila  &85 / 84 &80 / 80 \\
  &&RAIN &72 / 73  & 55 / 54 \\
&&\textbf{ours} &\textbf{ 61 / 64} & \textbf{49 / 51}\\

  \cline{2-5}
  &\multirow{3}*{\rotatebox{90}{\textbf{13b}}}
  &Vanila  &83 / 80 & 79 / 75\\
  &&RAIN & 38 / 39 & 32 / 30 \\
&&\textbf{ours} &\textbf{29 / 31} & \textbf{26 / 27}\\
\toprule

\end{tabular}

 \caption{Attack Success Rate. Under White-box Attacks (WBA.); Under Transferred Attacks (TA.). }
\label{tab:robust}
\end{table}

\subsubsection{Ablation Study. }We further analyze the impact of three types of uncertainties, namely semantic, factual, and value alignment, on the screening effectiveness to demonstrate the validity of the three dimensions we have defined. During the screening phase, we separately remove one of these dimensions for data control and observe the training results, as shown in Table \ref{tab:ablation}. For harmlessness, helpfulness tasks, and controlled emotion tasks, we choose the results of experiments on LLaMA 7B; for the truthfulness task, we choose the truth + info score. Results show that the three types of uncertainty signals each play a role in screening and exhibit significant complementarity. The full model that integrates multi-dimensional signals performs best, indicating that multi-dimensional modeling is an effective reliability assurance strategy. Additionally, the ablation study on the Phased Scheduling Training Strategy is presented in Table \ref{tab:strategy_analysis} of the preliminary experiment.

\begin{table}[h]
    \centering
  
  \begin{tabular}{ccccc}
    \toprule  
   \multirow{1}*{ \textbf{Uncertainty}}  
    &\multicolumn{1}{c}{ \textbf{Harm}}

    &\multirow{1}*{\textbf{Help}}
    &\multirow{1}*{\textbf{True}}
    &\multirow{1}*{\textbf{CS.}}
    
 \\
  \toprule
  
  w/o. $u_{sem}$& 6.7& 6.3 &67.8&75.6 \\

  w/o. $u_{fact}$&6.6&6.7&67.4&76.7\\
  w/o. $u_{align}$&6.7&6.6&69.5&77.5\\
\textbf{Full}&\textbf{7.0} &\textbf{7.2}& \textbf{74.2} &\textbf{82.6}\\
    
    \toprule
\end{tabular}  
 \caption{Ablation Study. We take the average score of the GPT-4 and Human evaluation. CS. represents controlled sentiment generation task. }
\label{tab:ablation}
\end{table}

\section{Conclusion}

This paper proposes an Uncertainty-Driven Adaptive Self-Alignment (UDASA) framework designed to enhance the automated alignment of LLMs. UDASA first generates multiple responses for the same input and quantifies their uncertainty across semantic, factual, and value dimensions. It then constructs preference pairs based on uncertainty differences and categorizes training samples into conservative, moderate, and exploratory phases to progressively guide model optimization. Experiments across harmlessness, helpfulness, factuality, and sentiment-controlled generation tasks demonstrate that UDASA significantly outperforms existing methods, improving model safety, effectiveness, and generalization.

%\bigskip

\bibliography{main}

\end{document}